\documentclass[conference]{IEEEtran}
\IEEEoverridecommandlockouts
\usepackage{cite}
\usepackage{amsmath,amssymb,amsfonts}
\usepackage{algorithmic}
\usepackage{graphicx}
\usepackage{textcomp}
\usepackage{xcolor}

\usepackage{algorithm}
\usepackage{multirow}%
\usepackage{booktabs}

\def\BibTeX{{\rm B\kern-.05em{\sc i\kern-.025em b}\kern-.08em
    T\kern-.1667em\lower.7ex\hbox{E}\kern-.125emX}}

\makeatletter
\newcommand{\linebreakand}{%
  \end{@IEEEauthorhalign}
  \hfill\mbox{}\par
  \mbox{}\hfill\begin{@IEEEauthorhalign}
}
\makeatother

\begin{document}

\title{Pruning for Sparse Diffusion Models Based on Gradient Flow}

\author{\IEEEauthorblockN{Ben Wan}
\IEEEauthorblockA{\textit{Shanghai Jiao Tong University}\\
Shanghai, China \\
burn-w@sjtu.edu.cn}
\and
\IEEEauthorblockN{Tianyi Zheng}
\IEEEauthorblockA{\textit{Shanghai Jiao Tong University}\\
Shanghai, China\\
tyzheng@sjtu.edu.cn}
\and
\IEEEauthorblockN{Zhaoyu Chen}
\IEEEauthorblockA{\textit{Fudan University}\\
Shanghai, China\\
zhaoyuchen20@fudan.edu.cn}
%\and
\linebreakand 
\IEEEauthorblockN{Yuxiao Wang}
\IEEEauthorblockA{\textit{Shanghai Jiao Tong University}\\
Shanghai, China\\
wyx980725@sjtu.edu.cn}
\and
\IEEEauthorblockN{Jia Wang$^{\ast}$ \thanks{*Corresponding author}}
\IEEEauthorblockA{\textit{Shanghai Jiao Tong University}\\
Shanghai, China\\
jiawang@sjtu.edu.cn}
}

\maketitle

\begin{abstract}
Diffusion Models (DMs) have impressive capabilities among generation models, but are limited to slower inference speeds and higher computational costs. Previous works utilize one-shot structure pruning to derive lightweight DMs from pre-trained ones, but this approach often leads to a significant drop in generation quality and may result in the removal of crucial weights. Thus we propose a iterative pruning method based on gradient flow, including the gradient flow pruning process and the gradient flow pruning criterion. We employ a progressive soft pruning strategy to maintain the continuity of the mask matrix and guide it along the gradient flow of the energy function based on the pruning criterion in sparse space, thereby avoiding the sudden information loss typically caused by one-shot pruning. Gradient-flow based criterion prune parameters whose removal increases the gradient norm of loss function and can enable fast convergence for a pruned model in iterative pruning stage. Our extensive experiments on widely used datasets demonstrate that our method achieves superior performance in efficiency and consistency with pre-trained models.
\end{abstract}

\begin{IEEEkeywords}
Diffusion Model, Structure Prune, Soft Prune, Progressive Prune, Gradient Flow
\end{IEEEkeywords}

\section{Introduction}
Diffusion models (DMs)~\cite{b1,b2,b3} have gained widespread attention in the field of generative modeling in recent years. The core idea is to generate high-quality samples by progressively adding noise to disrupt the data distribution and then gradually restoring the data through learning the reverse diffusion process. These models excel in tasks such as image generation~\cite{b4,b5}, audio synthesis~\cite{b6,b7}and text generation~\cite{b8,b9}, particularly in terms of the clarity and diversity of the generated images. However slower inference speeds and higher computational costs~\cite{b10,b11,b12} make DMs difficult to applied at resource-constrained mobile devices.

To reduce the high computational costs, significant efforts have been made to improve DMs by focusing on both training and sampling. Improvements in training include enhancing network structures~\cite{b10}, optimizing training strategies~\cite{b2}, and applying model distillation~\cite{b14,w1}. On the sampling side, efficient sampling strategies have been adopted to accelerate the process~\cite{b15,b12}. Although these methods reduce the consumption of computational resources and speed up inference, they often require retraining of the DMs and do not take full advantage of pre-trained models.

To tackle this issue, some pruning methods have been proposed for DMs. Diff-pruning~\cite{b16} proposes a structured pruning method, where one-shot pruning based on the Taylor pruning criterion is performed for the pre-trained DMs, followed by fine-tuning. However, the one-shot method tends to make the accuracy of the model drop a lot at once, and the more important weights may be clipped. Therefore, SparseDM~\cite{b17} proposes a progressive sparse fine-tuning method, which enables the pruning process to be continuous and smooth. Nevertheless, the consumption of both time and computational resources is large when using FID loss as pruning criterion.

In this paper, we propose a pruning method based on gradient flow~\cite{b18}, including the gradient flow pruning process and the gradient flow pruning criterion~\cite{b19}. Firstly the method adopts soft pruning strategy~\cite{b20,b21}, where the mask matrix is continuous rather than binary. Second, the method employs progressive prune strategy, which makes the sparsity of the model a continuous and smooth transition before one-shot pruning, reducing the information loss caused by pruning. Above two strategies make the whole prune process achieve continuity along gradient flow of energy function based on the pruning criterion in the mask space. Finally, our method adopts the gradient flow pruning criterion~\cite{b19}. Compared to loss-preservation based measures, gradient-norm based measures prune parameters whose removal increases the gradient norm and can enable fast convergence for a pruned model~\cite{b23}. In progressive soft prune process, those parameters that decrease the gradient flow should also be pruned, while those that increase the gradient flow should be retained. Our contributions can be summarized as follows:
\begin{itemize}
    \item We adopt soft pruning and progressive prune, which make the whole prune process achieve continuity along gradient flow in the mask space and avoid excessive degradation in generation quality by one-shot pruning.
    \item We utilize gradient flow pruning criterion in progressive soft prune process, which can enable fast convergence for a pruned model.
    \item We conduct experiments on image generation across multiple datasets, achieving competitive performance compared to other pruning methods. Extensive ablation studies are conducted to validate the effectiveness of our proposed method.
\end{itemize}

\section{Methodology}

% \subsection{Overview}

\subsection{Diffusion Models}

% Given a data distribution $q(\boldsymbol{x})$, diffusion models generate a distribution $p_{\theta}(\boldsymbol{x})$ to approximate $q(\boldsymbol{x})$, taking the form:
% \begin{equation}
% \begin{aligned}
% p_\theta(\boldsymbol{x}) &=\int p_\theta\left(\boldsymbol{x}_{0: T}\right) d \boldsymbol{x}_{1: T}\\
% &= \int p\left(\boldsymbol{x}_T\right) \prod_{t=1}^T p_\theta\left(\boldsymbol{x}_{t-1} \mid \boldsymbol{x}_t\right) d \boldsymbol{x}_{1: T}
% \end{aligned}
% \end{equation}
% where $\boldsymbol{x}$ is latent variables and the joint distribution $p_\theta\left(\boldsymbol{x}_{0: T}\right)$ can be expressed by learned Gaussian transitions $p_\theta\left(\boldsymbol{x}_{t-1} \mid \boldsymbol{x}_t\right)=\mathcal{N}\left(\boldsymbol{x}_{t-1} ; \mu_\theta\left(\boldsymbol{x}_t, t\right), \Sigma_\theta\left(\boldsymbol{x}_t, t\right)\right)$.

The diffusion model~\cite{b1,b2,b3,w2,w3,w4} consists of a forward process and a backward process. In the forward process, noise is incrementally added to the original image $q(\boldsymbol{x})$, generating a noisy image through a Markov chain, guided by a predefined variance schedule $\beta_{1: T}$. The backward process involves removing noise from the noisy image to restore the original image as accurately as possible. The training process of DMs is the process of establishing noise prediction models. The loss function of DM is minimizing a noise prediction objective:
\begin{equation}
\mathcal{L}(\boldsymbol{\theta}):=\mathbb{E}_{t, \boldsymbol{x}_0 \sim q(\boldsymbol{x})}\left[\left\|\boldsymbol{\epsilon}-\boldsymbol{\epsilon}_{\boldsymbol{\theta}}\left(\sqrt{\bar{\alpha}_t} \boldsymbol{x}_0+\sqrt{1-\bar{\alpha}_t} \boldsymbol{\epsilon}, t\right)\right\|^2\right],
\label{loss}
\end{equation}
where $\boldsymbol{\epsilon} \sim \mathcal{N}(0,1)$, $\alpha_{t}=1-\beta_{t}$ and $\bar{\alpha}_t=\prod_{s=1}^t \alpha_s$. After training, images can be sampled through the backward process from a noise.

% After training, images $\boldsymbol{x}_0$ can be sampled through an iterative process from a noise $\boldsymbol{x}_T\sim \mathcal{N}(0,1)$:
% \begin{equation}
% \boldsymbol{x}_{t-1}=\frac{1}{\sqrt{\alpha_t}}\left(\boldsymbol{x}_t-\frac{\beta_t}{\sqrt{1-\bar{\alpha}_t}} \boldsymbol{\epsilon}_{\boldsymbol{\theta}}\left(\boldsymbol{x}_t, t\right)\right)+\sigma_t \boldsymbol{z},
% \end{equation}
% where $\boldsymbol{z} \sim \mathcal{N}(\mathbf{0}, \boldsymbol{I})$ for $t>1$ and $\boldsymbol{z}=0$ for $t=1$. In this work, we aim to craft a lightweight $\boldsymbol{\epsilon}_{\boldsymbol{\widetilde{\theta}}}$ by removing redundant parameters of $\boldsymbol{\epsilon}_{\boldsymbol{\theta}}$, which are expected to produce similar $\boldsymbol{x}_0$ while the same $\boldsymbol{x}_T$ are presented.

\subsection{Soft Prune}

% This section we introduce what is soft pruning and how to realize the contiguity in sparse space. 

Previous work~\cite{b16} utilizes one-shot pruning\cite{b24} to compress DMs and fine-tune the pruned models to improve the performance. The pruning is defined as:
\begin{equation}
\min _{\mathcal{M}\in \{0,1\}^{d_1\times d_2}} \mathcal{E}(\mathcal{M}) \triangleq \mathcal{E}\left(\boldsymbol{\theta} \odot \mathcal{M}\right) \text {, s.t. } 1-||\mathcal{M}||_{0}/d_1 d_2=s,
\end{equation}
where $\boldsymbol{\theta}\in \mathbb{R}^{d_1\times d_2}$ is the dense weight for noise prediction network, $\mathcal{E}$ is a energy function based on a pruning criterion , $s$ is target sparsity, $\mathcal{M}$ is a zero-one mask matrix and $\odot$ represents element-wise multiplication. 

However, one-shot pruning tends to make the generation quality of the model drop a lot at once~\cite{b17}. Although fine-tuning can bring the generation performance of the model back up, some information of the model is still lost and some important weights are removed, which make the fine-tuning performance is limited. Thus, inspired by iterative magnitude pruning (IMP)~\cite{b22,b25}, we combine soft pruning with iterative pruning to avoid information loss and preserve more important weights.

In soft pruning, $\mathcal{M}$ is no longer a zero-one mask matrix and is defined as follows:
% \begin{definition}[Soft Pruning]
% Given a reference model $\boldsymbol{\theta}\in \mathbb{R}^{d}$, a target sparsity $s\in [0,1]$, a smooth energy function $\mathcal{E}$ based on a pruning criterion and a soft sparsity function $G$, the soft pruning model is defined as:
\begin{equation}
\min _{\mathcal{M}\in \{p,1\}^{d_1\times d_2}} \mathcal{E}(\mathcal{M}) \triangleq \mathcal{E}\left(\boldsymbol{\theta} \odot \mathcal{M}\right) \text {, s.t. } G(\mathcal{M})=s,
\label{sp}
\end{equation}
where $p$ is a constant and $G(\mathcal{M})=1-||\mathcal{M}-p||_{0}/d_1 d_2$ is a soft sparsity function.
% \end{definition}

Compared with vanilla pruning, soft pruning has two difference features. The first is that vanilla pruning removes the pruned weights permanently while the soft-pruned weights can be updated in iterative soft pruning. The other one is the soft sparsity metric $G$ is a generalization of the conventional zero-norm sparsity in vanilla pruning. 

% Besides, we find that soft pruning is a specific case of Sparse Fine Straight-through Estimator (SR-STE)~\cite{b22}, whose the update formula is:
% \begin{equation}
% \boldsymbol{\theta}-\gamma(g+\lambda(\boldsymbol{\theta}-\widetilde{\boldsymbol{\theta}})),
% \end{equation}
% where $\gamma$ is learning rate, $g$ is the gradient of loss function, $\lambda$ is a constant and $\widetilde{\boldsymbol{\theta}}=\boldsymbol{\theta} \odot \mathcal{M}$ is the sparse weight after pruning. When $\lambda=1$ and $\mathcal{M}$ in $\widetilde{\boldsymbol{\theta}}$ is soft mask matrix, above equation becomes $\widetilde{\boldsymbol{\theta}}-\gamma g$, which is exactly our soft pruning strategy.

\subsection{Progressive Prune}

In conventional iterative pruning, the constant $p$ in Eq.\ref{sp} is set to zero. Although the situation of removing some important weights is avoided to some extent, there is still a certain loss of model information. Thus we adopt progressive pruning strategy\cite{b17}, where the constant $p$ in mask matrix and the target sparsity $s$ in Eq.\ref{sp} varies with the pruning iterations. The pruning model is modified as:
\begin{equation}
\min _{\mathcal{M}\in \{p_{t},1\}^{d_\times d_2}} \mathcal{E}(\mathcal{M}) \triangleq \mathcal{E}\left(\boldsymbol{\theta} \odot \mathcal{M}\right) \text {, s.t. } G(\mathcal{M})=s_{t}.
\label{pp}
\end{equation}

\subsection{Gradient Flow Pruning Criterion}

To avoid information loss brought by one-shot pruning, we adopt progressive soft pruning. During the progressive soft prune process, loss-preservation is not the primary consideration and we want the gradient flow preserved for fast convergence. We consider the gradient flow of the loss as Eq.\ref{loss} in each step to the first order:
\begin{equation}
\Delta \mathcal{L}_t(\boldsymbol{\theta})=\lim _{\epsilon \rightarrow 0} \frac{\mathcal{L}_{t}(\boldsymbol{\theta}+\epsilon \nabla \mathcal{L}_{t}(\boldsymbol{\theta}))-\mathcal{L}_{t}(\boldsymbol{\theta})}{\epsilon}=\nabla \mathcal{L}_{t}^{\top} \nabla \mathcal{L}_{t}.
\label{gra}
\end{equation}
The gradient flow of the pruned network is better to be preserved or even increased for fast convergence. Following GraSP~\cite{b19}, we add a perturbation and use a Taylor approximation to measure how removing a weight will affect the gradient flow after pruning:
\begin{equation}
\begin{aligned}
\mathbf{I}_{t}(\boldsymbol{\delta}) & =\Delta \mathcal{L}_{t}\left(\boldsymbol{\theta}_0+\boldsymbol{\delta}\right) - \Delta \mathcal{L}_{t}\left(\boldsymbol{\theta}_0\right) \\
& =2 \boldsymbol{\delta}^{\top} \nabla^2 \mathcal{L}_{t}\left(\boldsymbol{\theta}_0\right) \nabla \mathcal{L}_{t}\left(\boldsymbol{\theta}_0\right)+\mathcal{O}\left(\|\boldsymbol{\delta}\|_2^2\right) \\
& =2 \boldsymbol{\delta}^{\top} \mathbf{H}_{t} \mathbf{g}_{t}+\mathcal{O}\left(\|\boldsymbol{\delta}\|_2^2\right),
\end{aligned}
\end{equation}
where $\mathbf{I}$ measures the change to Eq.\ref{gra}, $\mathbf{H}$ and $\mathbf{g}$ is the hessian matrix and gradient of $\mathcal{L}_{t}$. If $\mathbf{I}$ is negative, then removing the corresponding weights will reduce the gradient ﬂow, while if it is positive, it will increase the gradient ﬂow. Therefore, to accelerate convergence, we like to remove the weights with lower $\mathbf{I}_t$. For each weight, the importance can be computed in following way:
\begin{equation}
\mathbf{I}_{t}(\boldsymbol{\theta}\odot \mathcal{M})=\boldsymbol{\theta} \odot \mathcal{M} \odot \mathbf{H}_{t} \mathbf{g}_{t}.
\label{imp}
\end{equation}
% Although this criterion is more commonly used for pruning models trained from scratch, we believe that in progressive soft prune process, those parameters that decrease the gradient flow should also be pruned, while those that increase the gradient flow should be retained.

\begin{figure*}[htpb]
\centering
\includegraphics[width=0.9\textwidth]{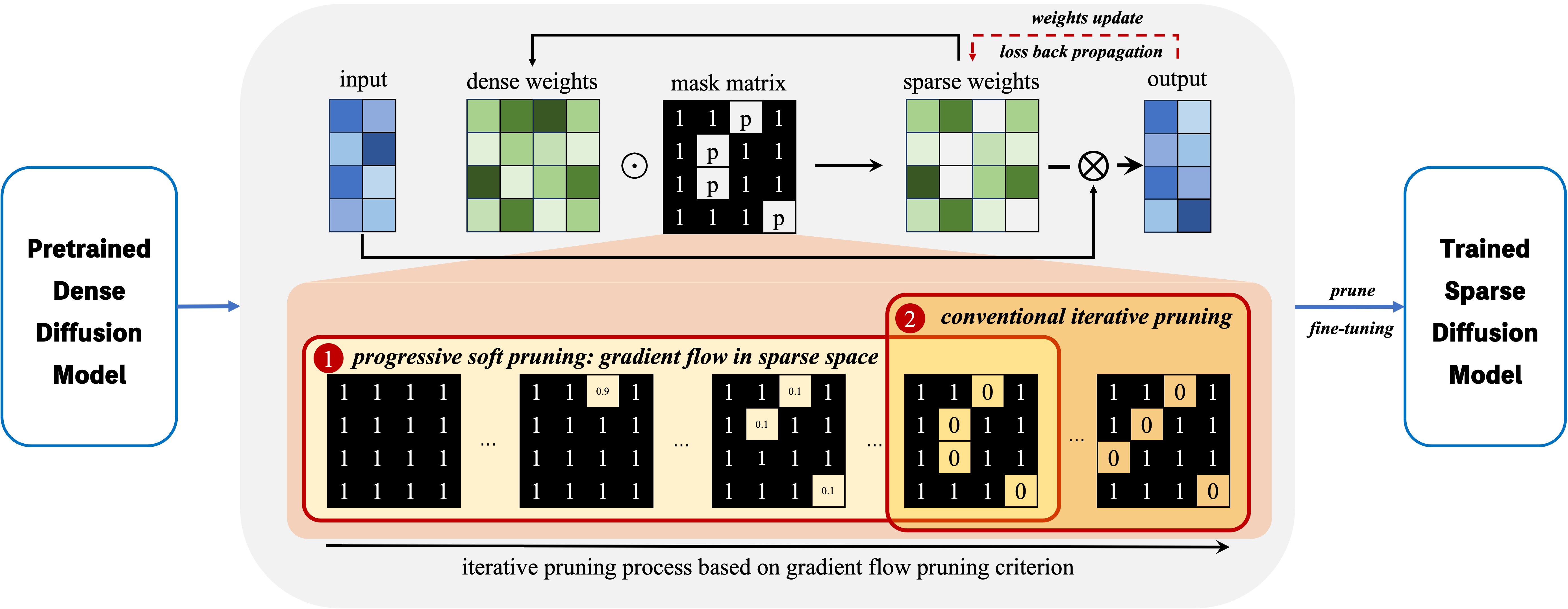}
\caption{Pruning for diffusion models based on gradient flow, including the gradient flow pruning process and the gradient flow pruning criterion.}
\label{fig}
\end{figure*}

\subsection{Gradient Flow in Sparse Space}

Fig.\ref{fig} shows our framework for DMs pruning. In iterative pruning process, each iteration can be divided into two steps, in which one is the update of the weights and the other is the update of the mask matrix. For convenience, each iteration we select the lowest importance using Eq.\ref{imp}, the corresponding elements of the mask matrix is set to $p_t$, and the rest is set to 1. Suppose the mask matrix $\mathcal{M}$ is belong to a sparse space. In previous work, the sparse space is discrete (e.g. $\{0,1\}^{d_1\times d_2}$). We utilize progressive soft pruning strategies to make the sparse space continuous (e.g. $\mathbb{R}^d$). According to the solution of mask matrix in Eq.\ref{pp}, optimization path of $\mathcal{M}$ can be regarded as the gradient flow in sparse space and each optimization step of $\mathcal{M}$ in sparse space is towards minimal energy function based on gradient flow pruning criterion. Based on the framework of our approach, we derive the gradient flow of the energy function adopting progressive soft pruning strategy:
\begin{equation}
\Delta \mathcal{E}(\mathcal{M}) = \nabla \mathcal{E}^{\top} \nabla \mathcal{E} = ||1-p_t\left(1 - \mathbb{I}_{A_{s_t}}(\mathbf{I}_{t}/||\mathbf{I}_{t}||_{2}) \right)||_{2},
\end{equation}
where $\mathbb{I}$ is indicator function and $A_{s_t}$ is the value of top $(1-s_t)d$-th largest normalized importance. The gradient flow in sparse space makes $\mathcal{M}$ can find the better mask matrix.

To enhance performance, we apply conventional iterative pruning after progressive soft pruning, allowing the sparse models to recover important weights that were previously clipped. The whole framework is as Algorithm.\ref{alg}. Both hyperparameter $p_t$ in mask matrix and current sparsity $s_t$ in Eq.\ref{pp} are linearly transformed with iterative steps $t$, where $p_t$ and $s_t$ is set to one and zero initially. When progressive soft pruning is finished, $p$ and $s_t$ turn to be zero and target sparsity $s$.

\begin{algorithm}[htpb]
\caption{Diffusion Models with Progressive Soft Pruning}\label{alg}
\textbf{Input:} A pre-trained DM $\boldsymbol{\theta}$, a dataset $X$, a target sparsity $s$, total training iterations $K$, iterative pruning iterations $M$ and progressive soft pruning iterations $N$ \\
\textbf{Output:} A pruned DM $\widetilde{\boldsymbol{\theta}}$
\begin{algorithmic}[1]
\STATE Initial $\widetilde{\boldsymbol{\theta}}=\boldsymbol{\theta}$
\FOR {$t$ in $M$:}
    \IF {$t<N$}
        \STATE Calculate current sparsity $s_t=ts/N$ and the mask value $p_t=1-t/N$ in mask matrix
        \ELSE
        \STATE Let $s_t=s$ and $p_t=0$
        \ENDIF
    \STATE Soft prune $s_t$ of weights in $\widetilde{\boldsymbol{\theta}}$ with $p_t$ mask matrix based on the importance Eq.\ref{imp}
\ENDFOR
\STATE Prune $s$ of weights in $\widetilde{\boldsymbol{\theta}}$ based on Diff-Pruning
\STATE Finetune the pruned model $\widetilde{\boldsymbol{\theta}}$ on $X$ for $K-M$ iterations
\STATE \textbf{return} the pruned DM $\widetilde{\boldsymbol{\theta}}$
\end{algorithmic}
\end{algorithm}

\section{Experiments}

\subsection{Experiments Setups}

\textbf{Datesets.} We evaluate our methods on CIFAR-10~\cite{b26}, CelebA-HQ~\cite{b27} and LSUN~\cite{b28}. Following previous works, we use the number of parameters (\#Params) and Multiply-Add Accumulation (MACs) as efficiency metrics, \textit{Frechet inception distance} (FID)~\cite{b29} as quality metric and Structural Similarity (SSIM)~\cite{b30} as consistency metric to evaluate our method.

\textbf{Implementation Details.} We use Denoising Diffusion Probability Models (DDPMs)~\cite{b1} as DMs in our experiments and deploy a 100-step sampler. The target sparsity is set to $0.5$. The total training iterations $K$ in Algorithm.\ref{alg} are 100K for CIFAR-10 and CelebA-HQ, 200K for LSUN-Bedroom and 500K for LSUN-Church. The iterative pruning iterations $M$ and progressive soft pruning iterations $N$ are set to $20\%$ and $10\%$ of total training iterations $K$ respectively.

\subsection{Experiments Results}

Table.\ref{exp} presents a comparison of our approach with other one-shot pruning methods~\cite{b24} on DMs. Magnitude pruning~\cite{b31}, taylor pruning~\cite{b32} and diff pruning~\cite{b16} deploy one-shot pruning strategy with different importance measures. Our approach achieves the best result on the four datasets. Using same train steps and achieving same compression performance, the lower FID implies the higher quality of generated samples and the larger SSIM score means more consistent generation with pre-trained model. Both show that our method achieves the best performance, demonstrating the efficiency and effectiveness of progressive soft pruning strategy.

\begin{table*}[htpb]
\caption{Diffusion pruning on CIFAR-10, CelebA-HQ, LSUN Church and Lsun Bedroom.}
  \label{exp}
\centering
\begin{tabular}{lcccccccccc}
\toprule[1pt]
 \textbf{Method} & \textbf{\#Params}$\downarrow$ & \textbf{\#Macs}$\downarrow$ & \textbf{FID}$\downarrow$ & \textbf{SSIM}$\uparrow$ & \textbf{Train Steps}$\downarrow$ &  \textbf{\#Params}$\downarrow$ & \textbf{\#Macs}$\downarrow$ & \textbf{FID}$\downarrow$ & \textbf{SSIM}$\uparrow$ & \textbf{Train Steps}$\downarrow$  \\
 \midrule[0.5pt]
& \multicolumn{5}{c}{\textbf{CIFAR-10 32 $\times$ 32 (100 DDIM steps)}} & \multicolumn{5}{c}{\textbf{CelebA-HQ 64 $\times$ 64 (100 DDIM steps)}} \\
\midrule[0.5pt]
Pretrained & 35.7M & 6.1G & 4.19 & 1.000 & 800K & 78.1M & 23.9G & 6.28 & 1.000 & 500K  \\
\midrule[0.5pt]
Magnitude Pruning & \multirow{4}{*}{19.8M} & \multirow{4}{*}{3.4G} & 5.48 & 0.929 & 100K & \multirow{4}{*}{43.7M} & \multirow{4}{*}{13.3G} & 7.08 & 0.870 &  100K \\
Taylor Pruning & & & 5.56 & 0.928 & 100K & & & 6.64 & 0.880 & 100K   \\
Diff Pruning~\cite{b16} &  &  & 5.29 & 0.932 & 100K &  &  & 6.24 & 0.885 & 100K   \\
Ours &  &  & \textbf{5.18} & \textbf{0.940} & 100K &  &  & \textbf{6.21} & \textbf{0.887} & 100K \\
  \midrule[0.5pt]
 & \multicolumn{5}{c}{\textbf{LSUN-Church 256 $\times$ 256 (100 DDIM steps)}} & \multicolumn{5}{c}{\textbf{LSUN-Bedroom 256 $\times$ 256 (100 DDIM steps)}} \\
\midrule[0.5pt]
 Pretrained & 113.7M & 248.7G & 10.6 & 1.000 & 4.4M &  113.7M & 248.7G & 6.9 & 1.000 & 2.4M  \\
\midrule[0.5pt]
 Magnitude Pruning& \multirow{4}{*}{63.2M} & \multirow{4}{*}{138.8G} & 22.0 & 0.742 & 500K & \multirow{4}{*}{63.2M} & \multirow{4}{*}{138.8G} & 63.9 & 0.668 &  200K \\
 Taylor Pruning & & & 19.0 & 0.773 & 500K & & & 29.5 & 0.749 & 200K   \\
 Diff Pruning~\cite{b16} & & & 13.9 & 0.762 & 500K & & & 18.6 & 0.764 & 200K   \\
 Ours & & & \textbf{12.5} & \textbf{0.791} & 500K & & & \textbf{17.9} & \textbf{0.772} & 200K   \\
\bottomrule[1pt]
\end{tabular}
\end{table*}

\begin{table*}[htpb]
\caption{Ablation study of soft pruning, progressive pruning and gradient flow pruning criterion. $+$ means adding the pruning strategy.}
  \label{abl}
\centering
\begin{tabular}{llcclcc}
\toprule[1pt]
 \textbf{Method} & \textbf{Pruning Criterion} & \textbf{FID}$\downarrow$ & \textbf{SSIM}$\uparrow$ & \textbf{Pruning Criterion} & \textbf{FID}$\downarrow$ & \textbf{SSIM}$\uparrow$ \\
 \midrule[0.5pt]
& \multicolumn{3}{c}{\textbf{CIFAR-10 32 $\times$ 32 (100 DDIM steps)}} & \multicolumn{3}{c}{\textbf{CelebA-HQ 64 $\times$ 64 (100 DDIM steps)}} \\
\midrule[0.5pt]
\multirow{2}{*}{Iterative Pruning}  & Magnitude & 5.38 & 0.932 & Magnitude & 6.81 & 0.875\\
% \quad $+$soft pruning & & 5.37 & 0.932 & & 6.74 & 0.877\\
% \quad $+$progressive pruning & & 5.35 & 0.933 &  & 6.66 & 0.879 \\
% \quad $+$progressive soft pruning (ours) & & 5.41 & 0.931  & & 7.02 & 0.872\\
% \hline
 & Taylor & 5.35 & 0.933 & Taylor & 6.43 & 0.882\\
% \quad $+$soft pruning & & 5.34 & 0.933 & & 6.39 & 0.883\\
% \quad $+$progressive pruning & & 5.32 & 0.934 &  & 6.37 & 0.883 \\
% \quad $+$progressive soft pruning (ours) & & 5.45 & 0.930 & & 6.50 & 0.883\\
\midrule[0.5pt]
Iterative Pruning & \multirow{4}{*}{Gradient Flow} & 5.28 & 0.933 & \multirow{4}{*}{Gradient Flow} & 6.29 & 0.884\\
\quad $+$soft pruning & & 5.25 & 0.934 & & 6.24 & 0.885\\
\quad $+$progressive pruning &  & 5.25 & 0.935 & & 6.23 & 0.886 \\
\quad $+$progressive soft pruning (ours) &  & \textbf{5.18} & \textbf{0.940} & & \textbf{6.21} &  \textbf{0.887}\\
\bottomrule[1pt]
\end{tabular}
\end{table*}

\subsection{Ablation Study}

To further verify our approach, we conduct ablation study to explore the effect of soft pruning, progressive pruning and gradient flow pruning criterion in iterative pruning stage. 

In conventional iterative pruning process, the $p_t$ is set to zero in mask matrix and the current sparsity is always set to the target sparsity in each pruning iteration. Adding soft pruning means the hyperparameter $p_t$ is changed along the pruning iterations and adding progressive pruning means current sparsity $s_t$ is changed along the pruning iterations.

As shown in Table.\ref{abl}, the results of iterative pruning are better than one-shot pruning on average, indicating it's necessary to avoid sudden information loss typically caused by one-shot pruning. Besides, both the soft pruning strategy and the progressive pruning strategy can improve the performance of pruned model compared with vanilla model with iterative pruning. The best performance is achieved when we combine progressive pruning and soft pruning, demonstrating the effectiveness of the continuity in sparse space. 

% although removing soft pruning makes the performance of pruned model worse, it is still better than pruned model adopting one-shot pruning strategy even under different pruning criteria. Thus the soft pruning strategy can improve the performance of pruned model.

% Discarding progressive pruning means the current sparsity is always set to the target sparsity in each pruning iteration and the hyperparameter $p$ is still changed along the pruning iterations. As shown in Table.\ref{abl}, the results of removing progressive pruning are better than one-shot pruning on average, indicating it's necessary to avoid sudden information loss typically caused by hard pruning.

% Moreover, we conduct the experiments of removing soft pruning and progressive pruning. The results show that adding soft pruning or progressive pruning can improve the performance of pruned models compared with vanilla model with iterative pruning. The best performance is achieved when we combine progressive pruning and soft pruning, demonstrating the effectiveness of the continuity in sparse space. 

\begin{figure}[htpb]
\centering
\includegraphics[width=0.45\textwidth]{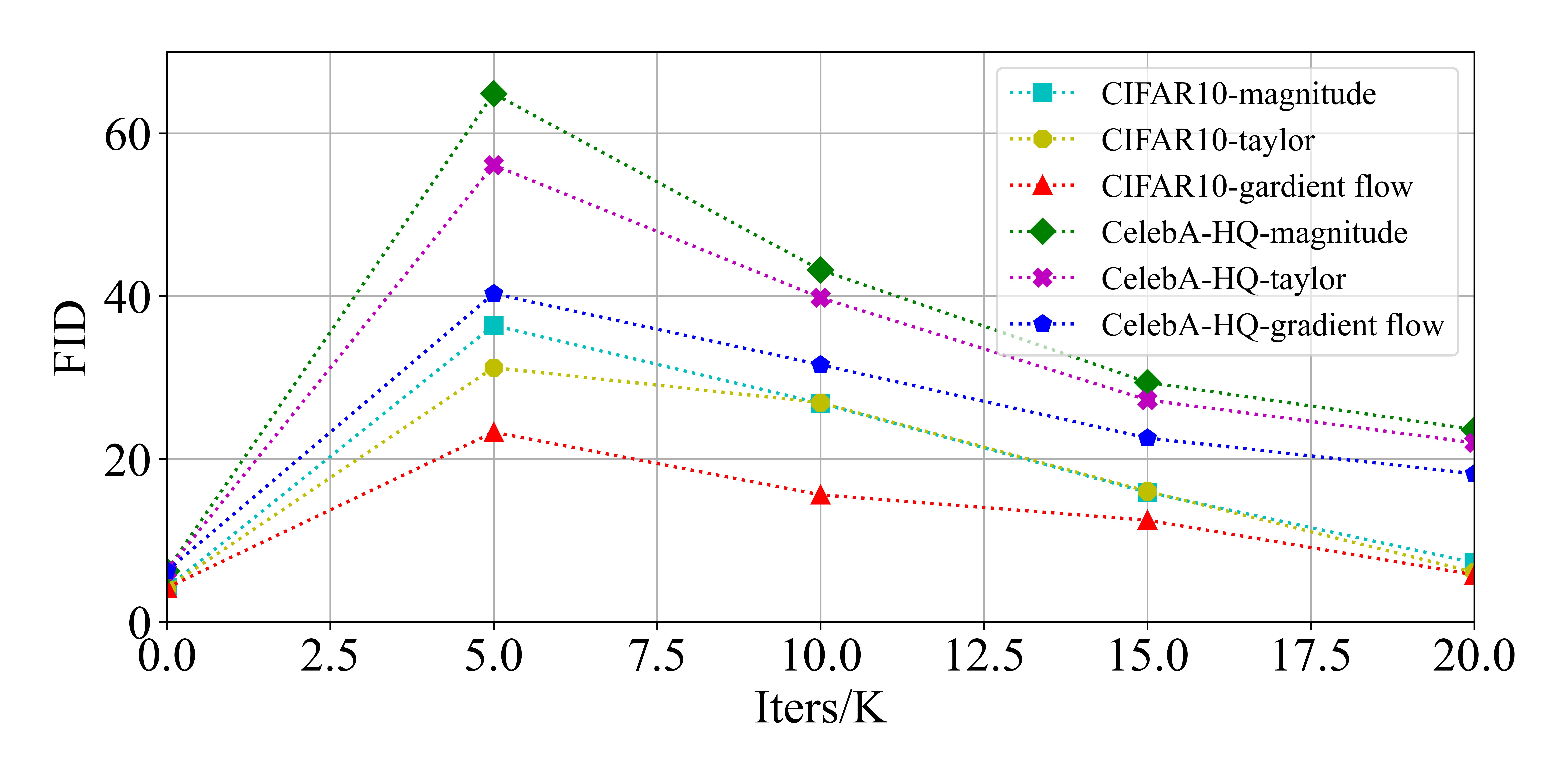}
\caption{FID change in iterative pruning stage using progressive soft pruning.}
\label{fig1}
\end{figure}

Moreover, adopting gradient flow pruning criterion in iterative stage outperforms other models adopting magnitude or taylor pruning criterion under same settings. To further verify that gradient flow pruning criterion can speed up the convergence of iterative pruning, we plot the figure of FID change in iterative pruning stage as Fig.\ref{fig1}. On both datasets, the iterative pruning adopting gradient flow pruning criterion can speed up the convergence of pruned model and attain lower FID, further demonstrating our approach effectiveness.

\section{Conclusion}

% In this paper, we propose an iterative pruning method based on gradient flow, incorporating a gradient flow pruning process and a gradient flow pruning criterion. Our approach utilizes a progressive soft pruning strategy to maintain the continuity of the mask matrix, guiding it along the gradient flow of the energy function based on pruning criterion in sparse space and preventing the abrupt information loss typically associated with one-shot pruning. The gradient-flow-based criterion prunes parameters whose removal increases the gradient norm of loss function, enabling faster convergence during the iterative pruning stage. Extensive experiments on widely used datasets demonstrate that our method achieves superior performance in both efficiency and consistency with pre-trained models, demonstrating our approach effectiveness.

In this paper, we propose an iterative pruning method based on gradient flow, incorporating both a pruning process and criterion. Our approach uses progressive soft pruning to maintain mask matrix continuity, guiding it along the gradient flow in sparse space and avoiding the abrupt information loss typical of one-shot pruning. The gradient-flow-based criterion prunes parameters that increase the gradient norm, enabling faster convergence during iterative pruning. Extensive experiments on widely used datasets show that our method outperforms in both efficiency and consistency with pre-trained models, highlighting its effectiveness.

\end{document}